\documentclass[letterpaper, 10 pt, conference]{ieeeconf}
\IEEEoverridecommandlockouts
\usepackage{cite}
\usepackage{amsmath,amssymb,amsfonts}
\usepackage{algorithm}
\usepackage{algcompatible}
\usepackage{graphicx}
\usepackage{textcomp}
\usepackage{xcolor}

\usepackage{amsthm}
\usepackage{acronym}
\usepackage{algpseudocode}
\usepackage{hyperref}
\usepackage{bm} %
\usepackage{array}

\newtheorem{theorem}{Theorem}[section]
\newtheorem{lemma}[theorem]{Lemma}

\newtheorem{definition}[theorem]{Definition}

\newenvironment{proofsketch}{\begin{proof}[Proof sketch]}{\end{proof}}

\acrodef{IMU}[IMU]{Inertial Measurement Unit}
\acrodef{PSR}[PSR]{Proprioceptively Stabilized Robot}
\acrodef{MPPI}[MPPI]{Model Predictive Path Integral}
\acrodef{COM}[COM]{Center of Mass}
\acrodef{GRU}[GRU]{Gated Recurrent Unit}
\acrodef{MLP}[MLP]{Multi Layer Perceptron}
\acrodef{UUB}[UUB]{Uniformly Ultimately Bounded}
\acrodef{CV}[CV]{Constant Velocity}

\allowdisplaybreaks
\begin{document}

\title{Learning Neural Observer-Predictor Models for \\Limb-level Sampling-based Locomotion Planning}
\author{Abhijeet M. Kulkarni$^{1}$, Ioannis Poulakakis$^{2}$ and Guoquan Huang$^{1}$
\thanks{$^{1}$A. M. Kulkarni and G. Huang are with the Department of Mechanical Engineering, University of Delaware, USA
        {\tt\small \{amkulk, ghuang\}@udel.edu}}%
\thanks{$^{2}$I. Poulakakis is with the School of Mechanical Engineering, National Technical University of Athens, Greece
        {\tt\small poulakas@mail.ntua.gr}}%
}

\maketitle

\begin{abstract}

Accurate full-body motion prediction is essential for the safe, autonomous navigation of legged robots, enabling critical capabilities like limb-level collision checking in cluttered environments. Simplified kinematic models often fail to capture the complex, closed-loop dynamics of the robot and its low-level controller, limiting their predictions to simple planar motion. To address this, we present a learning-based observer-predictor framework that accurately predicts this motion. Our method features a neural observer with provable \ac{UUB} guarantees that provides a reliable latent state estimate from a history of proprioceptive measurements. This stable estimate initializes a computationally efficient predictor, designed for the rapid, parallel evaluation of thousands of potential trajectories required by modern sampling-based planners. We validated the system by integrating our neural predictor into an \ac{MPPI}-based planner on a Vision 60 quadruped. Hardware experiments successfully demonstrated effective, limb-aware motion planning in a challenging, narrow passage and over small objects, highlighting our system's ability to provide a robust foundation for high-performance, collision-aware planning on dynamic robotic platforms.
\end{abstract}

\section{Introduction and Related Work}

Quadruped robots have progressed from laboratory prototypes to robust systems capable of real-world deployment. 
A key driver has been the maturity of low-level, proprioceptive controllers that fuse joint encoders and \acp{IMU} to maintain balance while tracking commanded high-level planar base velocities. 
We term robots with such onboard controllers as \acp{PSR} and this approach is now prevalent in both the research community \cite{Margolis2024IJRR, Norby2022ICRA} and on the commercial platforms \cite{Vision60, Unitree}. 
Consequently, the research focus in legged autonomy is shifting from the challenge of achieving dynamic locomotion to enabling safety-guaranteed autonomous navigation in complex environments like industrial inspection sites.

\ac{PSR} control permits the abstraction of complexity of legged locomotion into a simplified predictive model, such as omnidirectional kinematic models, for task-level planning. These models integrate seamlessly with classical navigation stacks and allow modular collision-aware planning \cite{Macenski2020IROS}. A typical high-level planner then selects velocity commands that accomplish a task in the presence of obstacles by combining perception with simplified dynamics. This paradigm was often adopted for legged \acp{PSR}, representing geometry with primitive shapes for collision checks \cite{Huang2023TRO, Narkhede2023IROS, Mattamala2022RAL, Li2024RAL, Liao2023IROS, Gaertner2021ICRA}.
\begin{figure}[t]
    \centering
    \includegraphics{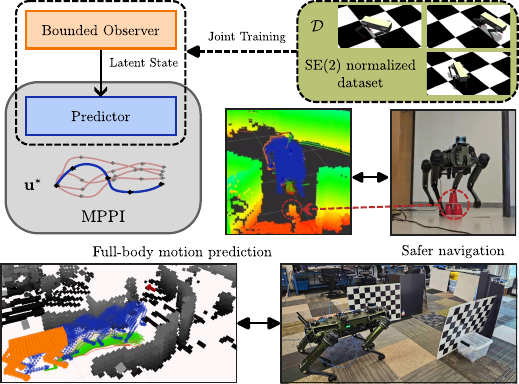}
    \caption{\textbf{The pipeline of the proposed system.} 
    The dataset is collected from both simulation and real robots and used to jointly train the proposed observer-predictor. The trained predictor is then deployed into the sampling-based MPPI planner for safe navigation. Full-body prediction allows for limb-level collision checking in narrow passages and over small objects.}
    \label{fig:overview}
\end{figure}

However, the effectiveness of these simplifications depends critically on how closely the actual \ac{PSR} response follows the assumed model under a given velocity command. Mismatch directly undermines the validity of collision checks and degrade task execution performance. To compensate, one can inflate the robot’s shape \cite{Gaertner2021ICRA} or employ robust formulations \cite{Lin2024Arxiv}. Data-driven alternatives predict the evolution of specific task related outputs conditioned on velocity commands—for example, forecasting \ac{COM} and footsteps \cite{Taouli2023IROS} or estimating collision probabilities \cite{Guzzi2020RAL,Kim2022RSS, Roth2025Arxiv}. Yet these approaches typically omit full-body motion prediction; moreover, collision probability estimators are often tied to a specific obstacle distribution and requires retraining whenever it changes.

There are works that are able to predict full-body motion based on motion library \cite{Zhang2018ACM} and even get planned commands \cite{Kim2023IROS}, but these are suitable for simulation environments where full state of the system is available. In case of real robotic system one does not have access to all states that are required for full-body motion prediction, and thus are to be recovered from measurements. Along with the states of physical body, \ac{PSR}s also have internal states associated with the low-level controller which are needed to be observed in order to predict the robot's motion. For such complex systems, learning based nonlinear stable observers exists \cite{Abdollahi2006NNLS, Chakrabarty2021Automatica, Jin2025NNLS}. By constraining the Lipschitz constant of the observer dynamics, one can establish stability guarantees \cite{Virmaux2018ANIPS}, a principle that has also supported stable learned controllers \cite{Shi2019ICRA}. What remains missing is a unified observer-predictor that recovers the latent state, produces full-body forecasts at planning rates, and slots into sampling-based, collision-aware planners without resorting to overly conservative geometry and simultaneously being agnostic to obstacle distribution.

As such, in this work, we design an efficient learning-based observer–predictor framework for full-body motion prediction of \acp{PSR},
see Fig.~\ref{fig:overview}, 
which is trained only from a history of proprioceptive observations and a sequence of future commands, and thus it is agnostic to obstacles.
Once trained, it can readily to be deployed on the PSR in {\em any} environments. 
In particular, the main contributions of the paper are the following:
\begin{itemize} 

\item  We design a neural observer that reconstructs the latent state of the \ac{PSR}, including the states of robot and low-level controller, from available measurements. 
We analytically derive conditions on the Lipschitz constant of the observer dynamics that ensure that it is \ac{UUB}. 

\item We build a neural predictor based on \ac{GRU}, which can perform massive parallel rollouts and enables thousands of trajectories to be evaluated concurrently for sampling-based planning. 
For instance, starting from a common initial latent state, the model rolls out 1000 command sequences for 200 steps (4 s with sampling rate of $\Delta t = 20$ms) in 13ms on a NVIDIA RTX T1000.

\item We integrate our neural observer/predictor with an \ac{MPPI} planner \cite{Williams2018TRO}, 
which can check limb-level collisions, avoid the conservatism of primitive shapes, and improve goal-pose tracking while retaining real-time performance.

\item We validate the proposed system in real settings, demonstrating accurate full-body prediction and safe navigation in cluttered environments,  outperforming the baseline \ac{CV} kinematic model.

\end{itemize}

\section{System Overview}

Consider the discrete-time evolution of a \ac{PSR} with a sampling period $\Delta t > 0$.
At each timestep $k$, the PSR receives a high-level planar (translational and rotational) velocity command as the input expressed in its body frame:
\begin{equation}\label{eq:user_commands}
    \mathbf u_k = \begin{bmatrix} v^{x}_k & v^{y}_k & \omega^{z}_k \end{bmatrix}^\top \in \mathbb{R}^{n_u}
\end{equation}
For collision-aware planning, the planner requires to predict the robot's full-body configuration, which includes its base position $\mathbf p_k \in \mathbb{R}^3$, orientation parameterized with roll-pitch-yaw $\bm \psi_k \in \mathbb{R}^3$, and joint angles $\bm \theta_k \in \mathbb{R}^{12}$. By convention, these quantities are expressed in a fixed, global reference frame.
However, the complete full-body configuration is not directly available from onboard sensors. The available proprioceptive measurements consist of the base's roll and pitch angles (which can be reliably estimated from the direction of gravity) and the joint angles. We thus define the directly measured output vector $\mathbf y_k$ and the full-body configuration vector $\mathbf z_k$ as follows:
\begin{align}
    \mathbf y_k =& \begin{bmatrix} \psi_{k}^\mathrm{roll} & \psi_{k}^\mathrm{pitch} & {\bm\theta}_k^\top \end{bmatrix}^\top \\
    \mathbf z_k =& \begin{bmatrix} \mathbf p_k^\top  & \psi_{k}^\mathrm{yaw} & \mathbf y_k^\top \end{bmatrix}
    \label{eq:observable_unobservable_outputs}
\end{align}

\subsection{Problem Statement}

A state estimator is typically used to infer the unmeasured components (global position and yaw) of $\mathbf z_k$, 
which however can be prone to drift over time if no global information is fused.
To mitigate the accumulation of global estimation errors, we instead predict future configurations \textit{relative} to the current robot's reference frame (akin to robocentric visual-inertial odometry~\cite{Huai2019IJRR}). 
Specifically, by modeling the PSR's dynamics with a latent state (implicit neural representation) $\mathbf x_k \in \mathbb{R}^{n_x}$, we seek to learn the mapping from historical data to a sequence of future relative configurations $\hat{\mathbf z}'_{k+1:k+T}$:
\begin{equation}
  \{  \underbrace{(\mathbf y_{k-H:k}, \mathbf u_{k-H:k})}_{\text{History}},  \underbrace{\mathbf u_{k+1:k+T}}_{\text{Future Commands}} \} ~\overset{\hat{\mathbf x}_k}{\Rightarrow}  \underbrace{\hat{\mathbf z}'_{k+1:k+T}}_{\text{Relative Predictions}}
\end{equation}

The core challenge is to recover an estimate $\hat{\mathbf x}_k$ of the latent state from the history and use it to  forecast the relative trajectory $\hat{\mathbf z}'_{k+1:k+T}$. 
These predictions are essential for a planner to perform limb-level collision checking against a local environment map \cite{Hornung2013AURO}. 
The non-smooth nature of this collision-checking problem--combined with the complex dynamics of the PSR--motivates the use of sampling-based planners such as \ac{MPPI}.
This choice, in turn, imposes a critical requirement on the learned model---that is, it has to be highly efficient, capable of generating thousands of trajectory rollouts in parallel to support real-time planning.

To that end, 
we learn a lightweight data-driven observer-predictor model 
and efficiently integrate it into a sampling-based planner.
Fig.~\ref{fig:overview} shows the overall pipeline of the proposed system that consists of three key stages:
(i) a nonlinear observer that recovers the latent state of the \ac{PSR} from a history of measurements (outputs) and motion commands (inputs),
(ii) a recurrent dynamics predictor that is initialized with this latent state and rolls out future full-body trajectories for candidate command sequences,
and (iii) a \ac{MPPI} planner that uses these predictions to select the optimal command by evaluating many potential outcomes in parallel.

\subsection{Decoupled Observer-Predictor Architecture}

\begin{figure}[t]
    \centering
    \includegraphics[width=\columnwidth]{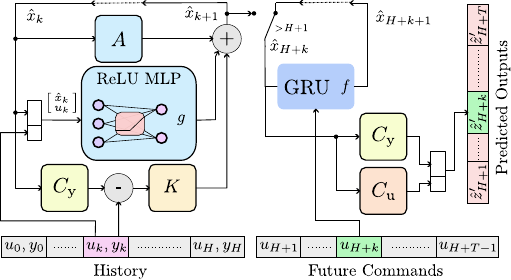}
    \vspace*{-5mm}
    \caption{\textbf{The proposed decoupled observer-predictor architecture.} 
    Our framework consists of a correction-based nonlinear observer (left) that produces a stable latent state estimate $\hat{\mathbf x}_k$ from history. This state initializes a recurrent predictor (right) that unrolls future relative full-body configurations~$\hat{\mathbf z}'$ in response to a candidate command sequence.}
    \label{fig:observer_predictor_arch}
\end{figure}

Fig.~\ref{fig:observer_predictor_arch} depicts the proposed  decoupled architecture of the state observer and the future dynamics predictor.  
This decoupling is crucial, as the observer's dynamics can be constrained to satisfy stability guarantees (see Sec.~\ref{sec:stability_of_nlo}), thus ensuring a reliable state estimate. 
The GRU-based predictor is optimized for the fast parallel rollouts required by the \ac{MPPI} planner.

In particular, a nonlinear observer is designed to recover the latent state $\hat{\mathbf x}_k$ from the history $(\mathbf u_{k-H:k}, \mathbf y_{k-H:k})$:
\begin{align}
    \hat{\mathbf x}_{k+1} &= \mathbf A\hat{\mathbf x}_k + g(\hat{\mathbf x}_k, \mathbf u_k) + \mathbf K(\mathbf y_k - \hat{\mathbf y}_k) \label{eq:observer}\\
    \hat{\mathbf y}_k &= \mathbf C_y \hat{\mathbf x}_k
\end{align}
where $\mathbf A\in \mathbb{R}^{n_x \times n_x}$ is a state transition matrix, 
$g(\cdot)$ is an \ac{MLP} network that models the underlying nonlinear dynamics, 
$\mathbf K\in \mathbb{R}^{n_x \times n_y}$ is the learned observer gain, 
and $\mathbf C_y\in \mathbb{R}^{n_y \times n_x}$ maps the state to measured outputs. 
It is important to note that this structure allows us to provide formal \ac{UUB} guarantees on the estimation error (see Thm.~\ref{thm:observer_stability_contraction}).

On the other hand, 
given the estimated state $\hat{\mathbf x}_k$, the predictor forecasts a sequence of relative full-body configurations $\hat{\mathbf z}'_{k+1:k+T}$ in response to a future command sequence $\mathbf u_{k+1:k+T}$. Starting with $\hat{\mathbf x}_k$, the prediction unrolls sequentially for $t=1,\dots,T$:
\begin{align} \label{eq:predictor}
    \hat{\mathbf x}_{k+t} =& f(\hat{\mathbf x}_{k+t-1}, \mathbf u_{k+t}) \\
    \hat{\mathbf z}'_{k+t} =& \begin{bmatrix} \mathbf C_\mathrm{u} \\ \mathbf C_y \end{bmatrix} \hat{\mathbf x}_{k+t}
\end{align}
where $f(\cdot)$ is an efficient GRU network that models the time evolution
and $\mathbf C_\mathrm{u} \in \mathbb{R}^{(n_z-n_y)\times n_x}$ maps the latent state to the unmeasured components of the relative configuration. The predictor's GRU structure was chosen for its effectieveness in modeling temporal sequences and its computational efficiency for the fast, parallel rollouts required by \ac{MPPI}.

\subsection{Joint End-to-End Training}

While functionally separated, the model parameters
\begin{equation}\label{eq:THETA}
    \bm \Theta = \{\mathbf A, \mathbf K, \mathbf C_y, g, f, \mathbf C_\mathrm{u}\} 
\end{equation}
of the observer and predictor are jointly trained on a quality dataset $\mathcal{D}$ that comprises diverse trajectories:
\begin{equation}
\mathcal{D} = \Big\{(\mathbf y_{0:H}^{(i)}, \mathbf u_{0:H+T}^{(i)}, \mathbf z'^{(i)}_{H+1:H+T})\Big\}_{i=1}^{N} \label{eq:dataset}
\end{equation}
To this end, the network is trained to minimize a loss function $\mathcal{L}(\Theta)$ that penalizes multi-step prediction errors while enforcing the observer's stability condition. 
This end-to-end training ensures the observer learns a latent representation that is both stable and maximally informative for the predictor's forecasting task.
Further details are provided in Sec.~\ref{sec:training}.

\subsection{MPPI Planning}\label{sec:mppi_prelim}

We integrate the learned predictor into an \ac{MPPI} framework \cite{Williams2018TRO} to select safe and task-aware commands. To improve sample efficiency, candidate command sequences $\mathbf u_{k+1:k+T}$ are parameterized by the control points $\{P_j\}$ of a degree-$d$ Bézier curve \cite{Higgins2023IROS}:
\begin{equation}\label{eq:bezier_curve}
    \mathbf u(s) = \sum_{j=0}^{d} \binom{d}{j} (1-s)^{d-j} s^j P_j, \qquad s \in [0,1].
\end{equation}
At each timestep, \ac{MPPI} samples noisy perturbations of the control points, generates corresponding command sequences, and simulates predicted trajectories using \eqref{eq:predictor}. The predicted relative trajectories $\hat{\mathbf z}'^{(i)}$ are converted to the global frame $\hat{\mathbf z}^{(i)}$ and evaluated by a cost function $J$ (detailed in Sec.~\ref{sec:mppi_sec}). The control points are then updated via a weighted average. The first command $\mathbf u_{k+1}^*$ of the optimized sequence is executed, and the process repeats. Alg.~\ref{alg:mppi_bezier} summarizes this process.

\begin{algorithm}[t]
    \caption{MPPI with Learned Predictor}
    \label{alg:mppi_bezier}
    \begin{algorithmic}[1]
        \State \textbf{Input:} Current latent state $\hat{\mathbf x}_k$, nominal control points $P$
        \For{$i=1,\dots,N$} \Comment{Sample N trajectories}
            \State Sample perturbed control points $P^{(i)} \sim \mathcal{N}(P, \Sigma)$
            \State Generate $\mathbf u^{(i)}_{k+1:k+T}$ from $P^{(i)}$ using \eqref{eq:bezier_curve}
            \State Predict $\hat{\mathbf z}'^{(i)}_{k+1:k+T}$ using \eqref{eq:predictor} from $\hat{\mathbf x}_k$
            \State Convert $\hat{\mathbf z}'^{(i)}$ to global $\hat{\mathbf z}^{(i)}$ using current robot pose
            \State Evaluate cost $\mathcal J^{(i)}$ \eqref{eq:total_cost_goal} with $\hat{\mathbf z}^{(i)}_{k+1:k+T}$ and $\mathbf u^{(i)}_{k+1:k+T}$
        \EndFor
        \State Compute weights $w^{(i)} \propto \exp(-J^{(i)}/\lambda)$
        \State Update control points $P \gets \sum_i w^{(i)} P^{(i)}$
        \State Execute the first command in the new optimal sequence
    \end{algorithmic}
\end{algorithm}

\section{Learning Observer and Predictor}
\label{sec:training}

In this section, we present in detail the process of learning our  neural observer and predictor. 
Note that the \ac{PSR} under consideration is the Ghost Robotics Vision60 quadruped~\cite{Vision60} equipped with its proprietary low-level controller that translates high-level velocity commands into joint torques.

\subsection{Building Datasets}
\label{sec:data_collection}

To build quality datasets that capture a rich variety of dynamic behaviors, including the complex internal modes of the low-level controller that may activate during navigation, we have collected 40-minute-long trajectories of the robot executing various commands in simulation and 10-minutes on the real robot in a motion capture environment. 
In simulations, the inertia parameters are randomized and noise is introduced in the measurements.
For each run $i$, we logged the command input $\mathbf u_k$, joint angles $\bm \theta_k$, and the ground-truth global base pose $(\mathbf p_k, \bm\psi_k)$ at a frequency of $1/\Delta t = 50~\mathrm{Hz}$ resulting in $N_\mathrm{dur}$ timesteps. This results in a raw dataset $\mathcal{D}_\mathrm{raw}^{(i)}=\{(\mathbf u_k, \mathbf p_k, \bm\psi_k, \bm\theta_k)\}_{k=0}^{N_\mathrm{dur}}$. To ensure the command distribution in our training data mirrors that of the downstream MPPI planner, we generated command sequences using piecewise Bézier segments. At the start of each segment, we sampled a duration $t_\mathrm{bez}\sim\mathsf{U}[3,7]~\mathrm{sec}$ and a set of control points $P_j\sim\mathsf{U}([-0.5,-0.5,-0.5],[0.5,0.5,0.5])$. The continuous command profile was then generated via \eqref{eq:bezier_curve} and discretized at $\Delta t$. 
This strategy produces smooth, non-stationary commands that effectively stimulate the range of the PSR's closed-loop dynamics possible during navigation.

From these collected raw logs, we extract overlapping windows of length $H+T+1$ (say $H=30$ and $T=200$). 
Each window forms a single data sample comprising an observation history $(\mathbf y_{0:H}, \mathbf u_{0:H})$, 
a future command sequence $\mathbf u_{H:H+T}$, 
and a ground-truth future full-body trajectory $\mathbf z_{H+1:H+T}$.
To ensure the model learns motions relative to the robot's current state rather than the global frame, we transform to robocentric coordinates.
This process can improve the sample efficiency and generalization by making learning invariant to the robot's initial state.
For each sample, we define a reference frame at the end of the history window, timestep $H$, with pose $(\mathbf p_H, \bm\psi_H)$. 
The future full-body configurations $\mathbf z_k$ for $k > H$ are transformed into this local frame to yield the relative configurations $\mathbf z'_k$. This transformation is defined as:
\begin{equation}
    \mathbf z'_k = \mathcal{T}_{(\mathbf p_H, \bm\psi_H)}^{-1} (\mathbf z_k) = 
    \begin{bmatrix}
        \mathbf R_z(-\psi_H^{\mathrm{yaw}}) (\mathbf p_k - \mathbf p_H) \\
        \psi_k^{\mathrm{yaw}} - \psi_H^{\mathrm{yaw}} \\
        \psi_k^{\mathrm{roll}} \\
        \psi_k^{\mathrm{pitch}} \\
        \bm\theta_k
    \end{bmatrix}
\end{equation}
where $\mathbf R_z(-\psi_H^{\mathrm{yaw}})$ is the rotation matrix for the yaw angle. 
Note that roll, pitch, and joint angles are already expressed in the body frame and thus remain unchanged. 
Applying this transformation process to all extracted windows yields the final processed dataset $\mathcal{D}$ as defined in \eqref{eq:dataset}.

\subsection{Loss Function} %

To train the model parameters $\bm\Theta$~\eqref{eq:THETA} of the observer and predictor,
we formulate the following composite loss function that combines a prediction accuracy term with a regularization term to enforce observer stability.

The primary objective is to accurately forecast future trajectories. 
For each sample in $\mathcal{D}$~\eqref{eq:dataset}, the observer \eqref{eq:observer} is first unrolled over the history $(\mathbf y_{0:H}, \mathbf u_{0:H})$ to compute the latent state estimate $\hat{\mathbf x}_H$. This state initializes the predictor~\eqref{eq:predictor}, which is then rolled out for $T$ steps using the future commands $\mathbf u_{H:H+T-1}$ to predict the relative trajectory $\hat{\mathbf z}'_{H+1:H+T}$. 
The prediction loss measures the mean squared error between the predicted and ground-truth relative trajectories:
\begin{equation}\label{eq:prediction_loss}
    \mathcal{L}_\mathrm{pred} = \frac{1}{T} \sum_{t=1}^{T} \big\| \mathbf z'_{H+t} - \hat{\mathbf z}'_{H+t} \big\|_2^2
\end{equation}
To ensure the observer is well-behaved and its state estimate $\hat{\mathbf x}_k$ converges towards the true state $\mathbf x_k$, we enforce the contraction condition from Thm.~\ref{thm:observer_stability_contraction}. This is achieved by penalizing violations of the condition $\rho = \|\mathbf A - \mathbf K \mathbf C_y\|_2 + L_g < 1$. We formulate this as a hinge loss:
\begin{equation}\label{eq:stability_loss}
    \mathcal{L}_\mathrm{stab} = \max\big(0, \rho - (1 - \epsilon)\big)
\end{equation}
where $\epsilon > 0$ is a small margin (e.g., $10^{-4}$) for strict contraction. During training, $\|\mathbf A - \mathbf K \mathbf C_y\|_2$ is computed as the spectral norm of the matrix. The Lipschitz constant $L_g$ of the MLP $g$ is upper-bounded using Lemma~\ref{lem:relu_lips} as the product of the spectral norms of its weight matrices.

The models are then trained end-to-end by minimizing the weighted sum of the aforementioned prediction and stability losses over the dataset $\mathcal{D}$~\eqref{eq:dataset}:
\begin{equation}
    \mathcal{L}(\Theta) = \mathbb{E}_{\mathcal{D}} \Big[ \mathcal{L}_\mathrm{pred} + \alpha \mathcal{L}_\mathrm{stab} \Big]
    \label{eq:total-loss}
\end{equation}
where $\alpha > 0$ is a hyperparameter balancing prediction accuracy with the observer stability guarantee. 
We parameterize the predictor's transition function $f$ as a \ac{GRU} network and use the Adam optimizer to find the optimal parameters $\bm\Theta^*$.
This joint training procedure yields an accurate predictor that is grounded in a provably stable state estimation framework.

\subsection{Model Analysis}

As the proposed neural observer-predictor model is the key enabler for efficient and safe mission-oriented MPPI-based navigation,
we perform numerical analysis to quantitatively understand its training and testing behaviors  along three key axes: 
(i) stability during training, (ii) generalization to unseen data, and (iii) prediction accuracy over a finite horizon. 
To this end, we consider the \ac{MLP}  in the observer has three layers of sizes $[128,128,128]$ and the predictor is a single layer GRU with hidden state of size $128$, 
which is equivalent to latent state of $n_x = 128$. 
The hyperparameter balancing prediction accuracy and observer stability  in \eqref{eq:total-loss} was set to $\alpha = 0.1$.

\begin{figure}[t]
    \centering
    \includegraphics{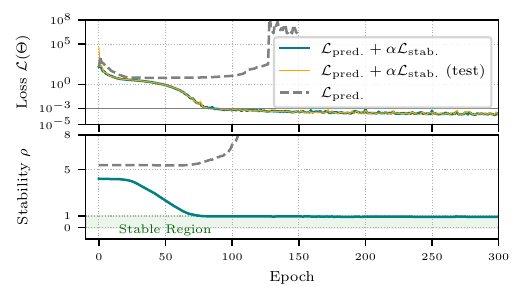}
    \vspace*{-10mm}
    \caption{\textbf{Training Behaviors.} Comparison of training with (blue) and without (orange) the stability regularization term \eqref{eq:stability_loss}. The stabilized model maintains the contraction factor $\rho$ strictly below unity, leading to smoother convergence and lower variance, while the unconstrained model quickly becomes unstable.}
    \label{fig:train_stability}
    \vspace*{-2mm}
\end{figure}

The significance of including the stability regularization~\eqref{eq:stability_loss} in the training is clearly demonstrated in Fig.~\ref{fig:train_stability}.
Without it, the observer parameters drift into the unstable regime ($\rho \geq 1$), causing the prediction error to diverge. In contrast, by incorporating the stability penalty, the contraction factor $\rho$ is constrained to converge strictly below unity. This not only guarantees the observer's contraction property but also leads to smoother loss curves and significantly lower variance across training runs. Importantly, stability enforcement also improves generalization. Across an $80/20$ train-test split, the test error of the stabilized model closely tracks the training error with no signs of overfitting. This suggests that the contraction property acts as an effective structural regularizer, preventing the model from learning unstable dynamics and enhancing its robustness to novel trajectories.

\begin{figure}[t]
    \centering
    \includegraphics{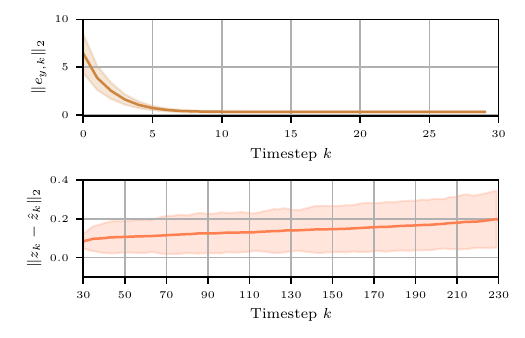}
    \vspace*{-8mm}
    \caption{\textbf{Test Performance.} 
    (top) Observer state estimation error converges rapidly from a large initial error distribution. 
    (bottom) The resulting predictor rollout error remains low and bounded over the full 4-second horizon, demonstrating accurate long-term forecasting.}
    \label{fig:observer_predictor_error} 
\end{figure}

Fig.~\ref{fig:observer_predictor_error} validates the performance of the two core models. 
The observer (top) consistently recovers the latent state, with the estimation error converging within 30 timesteps ($0.6$s) despite being initialized from a broad uniform distribution ($\hat{x}_0 \sim \mathsf{U}([-10,10]^{n_x})$). 
Once initialized with this stable state, the predictor (bottom) achieves low rollout error over the entire $T=200$ step horizon. 
This highlights the synergy between the two components: the observer provides a reliable latent state, enabling the predictor to accurately forecast future trajectories.

Interestingly, it can be seen from Fig.~\ref{fig:observer_predictor_error} that the empirical convergence rate is faster than the theoretical bound which  might suggest for $\rho \approx 0.97$ (from Fig.~\ref{fig:train_stability}). 
This is because our theoretical bound in Thm.~\ref{thm:observer_stability_contraction} is conservative, relying on worst-case upper estimates of the Lipschitz constants.

\section{MPPI Planing  for Goal-Pose Navigation}
\label{sec:mppi_sec}

To demonstrate its utility, we integrate our learned observer-predictor model into the \ac{MPPI} framework, enabling it to generate collision-aware trajectories for mission-oriented navigation.
The specific task is goal-pose tracking, where the objective is to drive the robot's planar pose, 
$\mathbf q_k=[p_k^x, p_k^y, \psi_k^\mathrm{yaw}]^\top$, to a user-defined goal, $\mathbf q_\mathrm{goal}$.
At each planning cycle $k$, the observer provides the latent state estimate $\hat{\mathbf x}_k$ from the recent measurement history. This state serves as the initial condition for the predictor, which generates $N$ future trajectory rollouts corresponding to perturbed command sequences $\mathbf u^{(i)}_{k+1:k+T}$. Each predicted relative trajectory, $\hat{\mathbf z}'^{(i)}_{k+1:k+T}$, is projected into the global frame using the robot's current pose $(\mathbf p_k, \bm\psi_k)$:
\begin{equation}
    \hat{\mathbf z}^{(i)}_{k+t} = \mathcal{T}_{(\mathbf p_k, \bm\psi_k)}(\hat{\mathbf z}'^{(i)}_{k+t}), \quad \forall  t=1,\dots,T.
\end{equation}
These global trajectories are then evaluated using a cost function $J^{(i)}$, detailed below. 
The nominal command is updated via the importance-sampling average, 
and the first command of the optimized sequence, $\mathbf u_{k+1}^*$, is dispatched to the robot.

\subsection{Cost Function}

\begin{figure}[t]
    \centering
    \includegraphics[width=\columnwidth]{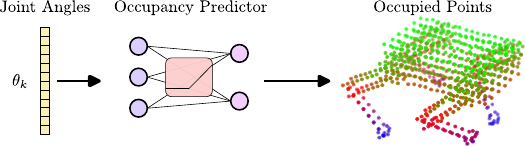}
    \vspace*{-8mm}
    \caption{\textbf{Robot Occupancy Predictor.} An ReLU MLP-based occupancy predictor, with layer sizes $[64,128]$, takes joint angles $\bm\theta_k$ as input and outputs a $780$ set of points ${}^{B}\mathcal{O}(\bm\theta_k)$ representing occupied space in the robot's base frame. For collision checking, these points are transformed to the global frame and compared against an environment map $\mathcal{M}$.}
    \label{fig:occupancy_predictor}
    \vspace*{-2mm}
\end{figure}

The cost function $J$ used in Alg.~\ref{alg:mppi_bezier} steers the MPPI planner by aggregating penalties for collision risk, goal tracking, and control effort over the prediction horizon $T$.
First of all, to ensure safety, we penalize trajectories that risk collision with the environment, which is represented as a voxel map $\mathcal{M}$. We use a separately trained\footnote{While its occupied volume can be computed via standard forward kinematics, a learned network offers a simpler, parallelizable implementation.} MLP, $h(\bm\theta)$, that maps joint angles to a point cloud ${}^{B}\!\mathcal{O}(\bm\theta)$ representing the robot's occupied volume in its base frame (see Fig.~\ref{fig:occupancy_predictor}). For each predicted state $\hat{\mathbf z}_{k+t}$ along a rollout, these body points are transformed into the global frame. The collision cost is the number of points that fall within occupied voxels:
\begin{equation}
\label{eq:collision_cost}
\ell_{\mathrm{coll}}(\hat{\mathbf z}_{k+t}) =
\sum_{\mathbf o \in {}^{B}\!\mathcal{O}(\hat{\bm\theta}_{k+t})}
\mathbf{1}_{\mathcal{M}}\!\big(\mathbf R_z(\hat{\bm\psi}_{k+t}) \mathbf o + \hat{\mathbf p}_{k+t}\big)
\end{equation}
where $\mathbf{1}_{\mathcal{M}}(\cdot)$ is an indicator function for the voxel map.

The task cost guides the robot towards a static goal configuration $\mathbf q^{\mathrm{goal}}$. It is composed of penalties on position and yaw.
The position cost at each timestep penalizes the Euclidean distance between the predicted 2D position $\hat{\mathbf p}^{xy}_{k+t}$ and the goal:
\begin{equation}
\ell_{\mathrm{pos}}(\hat{\mathbf z}_{k+t}) = \big\|\hat{\mathbf p}^{xy}_{k+t} - \mathbf p_{\mathrm{goal}}^{xy}\big\|_2
\end{equation}
The yaw cost dynamically transitions between aligning with the direction of travel when far from the goal and matching the final goal orientation when nearby. A desired heading, $\psi_{k+t}^\mathrm{head.}$, is determined as the direction of the velocity vector. A sigmoid weighting function $\sigma(\cdot)$, based on the distance to the goal, smoothly interpolates between these two sub-objectives:
\begin{equation}
\lambda = \sigma\left(\frac{\|\hat{\mathbf p}^{xy}_{k+t} - \mathbf p_{\mathrm{goal}}^{xy}\|_2 - d_{\mathrm{switch}}}{s_{\mathrm{switch}}}\right)
\end{equation}
where $d_{\mathrm{switch}}$ and $s_{\mathrm{switch}}$ control the transition's center and sharpness. The per-step yaw cost is a weighted average of the two alignment errors:
\begin{equation}
\ell_{\mathrm{yaw}}(\hat{\mathbf z}_{k+t}) = \lambda |\hat{\psi}^{\mathrm{yaw}}_{k+t} - \psi_{k+t}^\mathrm{head}| + (1-\lambda) |\hat{\psi}^{\mathrm{yaw}}_{k+t} - \psi_{\mathrm{goal}}^{\mathrm{yaw}}|
\end{equation}
The total goal-tracking cost at a timestep is $\ell_{\mathrm{goal},t} = \ell_{\mathrm{pos}} + \ell_{\mathrm{yaw}}$.
To promote smooth motions, the control cost adds a quadratic penalty on the magnitude of the control inputs:
\begin{equation}
\ell_{\mathrm{ctl}}(\mathbf u_{k+t}) = \|\mathbf u_{k+t}\|_2^2
\end{equation}

Finally, the total cost for a sampled trajectory is the weighted sum of these costs over the prediction horizon:
\begin{align}
\label{eq:total_cost_goal}
J^{(i)} = \sum_{t=1}^{T}\Big[&
\mathrm{w}_c\ell_{\mathrm{coll}}(\hat{\mathbf z}^{(i)}_{k+t})
+ \mathrm{w}_g\ell_{\mathrm{goal},t}(\hat{\mathbf z}^{(i)}_{k+t})
+ \mathrm{w}_u\ell_{\mathrm{ctl}}(\mathbf u^{(i)}_{k+t})
\Big]
\end{align}
where the positive weights $\mathrm{w}_c, \mathrm{w}_g$, and $\mathrm{w}_u$  balance safety, task performance, and control effort. This cost determines the weights used in the MPPI update rule (see Alg.~\ref{alg:mppi_bezier}).

\section{Stability Analysis of the Neural Observer} \label{sec:stability_of_nlo}

At this point, 
we have demonstrated the practical aspects of the proposed neural observer/predictor in goal-tracking legged navigation.
Equally significant from a theoretical standpoint, we establish the conditions under which our neural observer is guaranteed to be \ac{UUB}.
This guarantee is crucial, as it formally ensures that the latent state estimate remains close to the true value, 
thereby providing a reliable foundation for prediction and planning, regardless of applications.

Our analysis relies on a key common assumption about the true system dynamics 
and leverages the concept of Lipschitz continuity to handle the nonlinearities introduced by the neural network.
Specifically, we assume the true closed-loop dynamics of the PSR can be represented by the observer's learned structure, perturbed by a bounded disturbance term $\epsilon_k$. Specifically, the true latent state evolves according to:
\begin{equation}\label{eq:true_dynamics_for_proof}
    \mathbf x_{k+1} = \mathbf A \mathbf x_k + g(\mathbf x_k, \mathbf u_k) + \bm\epsilon_k
\end{equation}
where the disturbance is bounded, i.e., $\|\bm\epsilon_k\|_2 \leq \epsilon_{\max}$. 
This assumption is justified because the parameters of $\mathbf A$ and the MLP $g(\cdot)$ are learned to capture the dominant system dynamics from the training data $\mathcal{D}$, while $\bm\epsilon_k$ provides robustness to the inevitable imperfections of the learned model and noise.

\subsection{Lipschitz Continuity of MLP}

Our \ac{UUB} guarantee requires bounding the effect of $g(\cdot)$, 
which a \ac{MLP} network achieved with its Lipschitz constant.
\begin{definition}[Lipschitz Continuity \cite{Khalil2002PH}] \label{def:lipschitz}
A function $g: \mathbb{R}^n \rightarrow \mathbb{R}^m$ is globally $L_g$-Lipschitz continuous if there exists a constant $L_g > 0$ such that for all $\mathbf x, \mathbf y \in \mathbb{R}^n$, the inequality $\|g(\mathbf x) - g(\mathbf y)\|_2 \leq L_g\|\mathbf x - \mathbf  y\|_2$ holds.
\end{definition}

For a ReLU MLP as used in our model, the Lipschitz constant can be upper-bounded by the product of the spectral norms of its weight matrices based on the following lemma, which allows us to control $L_g$ during training:
\begin{lemma}[Lipschitz constant of ReLU \ac{MLP} \cite{Virmaux2018ANIPS}]\label{lem:relu_lips} 
Let $ g(\mathbf x)= \mathbf W^{N+1}\phi\bigl(\mathbf W^{N}(\cdots \phi(\mathbf W^{1} \mathbf x+\mathbf b^{1})\cdots)+\mathbf b^{N}\bigr), $ 
where each $\mathbf W^{n}$ is a matrix, $\mathbf b^{n}$ a bias vector, and $\phi(\mathbf x)=\max\{\mathbf x,0\}$ is applied element-wise. Then an upper bound on the Lipschitz constant of the ReLU \ac{MLP} is $L_g\le\prod_{n=1}^{N+1}\|\mathbf W^{n}\|_2$.
\end{lemma} 
\begin{proofsketch}
See Appendix~\ref{sec:proof_relu_lem}.
\end{proofsketch}

\subsection{UUB Guarantee via Contraction Analysis}

With these preliminaries, we now have the following key result, 
analytically establishing the UUB guarantee of the proposed neural observer based on a contraction condition:
\begin{theorem}[UUB of Observer Error]
\label{thm:observer_stability_contraction}
Consider the observer from Eq.~\eqref{eq:observer} and assume the true system dynamics are given by \eqref{eq:true_dynamics_for_proof} with a linear output $\mathbf y_k = \mathbf C_y \mathbf x_k$. 
Let the estimation error be $\mathbf e_k = \mathbf x_k - \hat{\mathbf x}_k$. 
Define the closed-loop error dynamics matrix as $\mathbf A_c = \mathbf A - \mathbf K \mathbf C_y$. 
If the spectral norm of $\mathbf A_c$ and the Lipschitz constant $L_g$ of the MLP $g$ satisfy the contraction condition
\begin{equation}\label{eq:contraction_condition}
    \rho := \|\mathbf A_c\|_2 + L_g < 1,
\end{equation}
then both the state estimation error $\mathbf e_k$ and the output estimation error $\mathbf e_{y,k} = \mathbf y_k - \hat{\mathbf y}_k$ are Uniformly Ultimately Bounded. 
Specifically, their norms are ultimately bounded by:
\begin{equation*}
    \limsup_{k \to \infty} \|\mathbf e_k\|_2 \le \frac{\epsilon_{\max}}{1 - \rho} ~~ \text{and} ~~ \limsup_{k \to \infty} \|\mathbf  e_{y,k}\|_2 \le \|\mathbf C_y\|_2 \frac{\epsilon_{\max}}{1 - \rho}.
\end{equation*}
\end{theorem}
\begin{proofsketch}
See Appendix~\ref{sec:proof_uub_thm}.
\end{proofsketch}

\section{Experimental Validation}

In this section, we present an experimental validation of our proposed observer-predictor neural models. We first perform a comparative analysis, evaluating our model's prediction accuracy and computational performance against a standard \ac{CV} baseline. We then demonstrate the effectiveness of the complete system by integrating our model into the \ac{MPPI} planner for a challenging goal-pose navigation task on the Vision 60 quadruped robot~\cite{Vision60}.

\subsection{Comparative Analysis with Baseline Model}
The \ac{CV} model predicts the future 2D pose given the current pose and commanded velocity:
\begin{equation} \label{eq:constant_vel_model}
    \mathbf q_{k+1} = \mathbf q_{k} + \mathbf R_z(\psi^\mathrm{yaw}_k) \mathbf u_k \Delta t
\end{equation}
where $\mathbf q\in \mathbb{R}^3$ is the 2D planar pose. While this model provides a strong baseline, it assumes perfect velocity tracking, an idealization that often fails in practice due to complex robot dynamics and low-level control interactions. We evaluate our proposed model against this baseline for prediction accuracy and computational efficiency.

\subsubsection{Prediction Accuracy}

\begin{figure}[t]
    \centering
    \includegraphics[width=\linewidth]{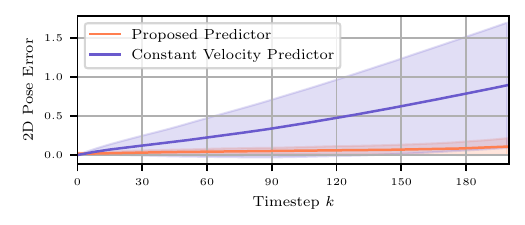}
    \vspace*{-10mm}
    \caption{\textbf{Long-Horizon Prediction Error.} Comparison of 2D pose prediction error over a 4-second horizon. Our proposed model (orange, mean with $\pm 1\sigma$ region) maintains low error. The baseline \ac{CV} model's error (dashed orange) accumulates rapidly, highlighting its unsuitability for long-horizon planning.}
    \label{fig:comp_cv_proposed_error} 
\end{figure}

As shown in Fig.~\ref{fig:comp_cv_proposed_error}, our model's prediction error remains low and bounded over a 4-second horizon. In contrast, the \ac{CV} model's error grows without bound. This confirms that our model successfully captures the \ac{PSR} dynamics, making it far more reliable for use in a predictive planner.

\subsubsection{Computational Efficiency}

\begin{figure}[t]
    \centering
    \includegraphics[width=\linewidth]{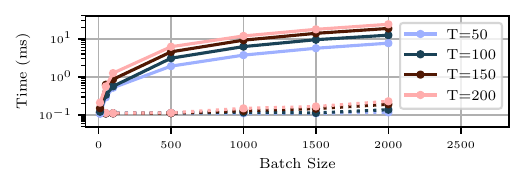}
    \vspace*{-10mm}
    \caption{\textbf{Inference Time Comparison.} Computation time versus batch size and prediction horizon for our model (solid lines) and the CV baseline (dashed lines) on an Nvidia RTX T1000 GPU. While more intensive, our model scales efficiently and remains well within real-time limits.} 
    \label{fig:comp_cv_proposed_times}
\end{figure}

For real-time planning, computational efficiency is paramount. We benchmarked GPU-accelerated C++ Torch implementations of both models (see Fig.~\ref{fig:comp_cv_proposed_times}). While our GRU-based predictor is naturally more complex than the simple CV model, its inference time scales efficiently with batch size and horizon length. The performance is well within the required limits for real-time sampling-based planners like \ac{MPPI}, demonstrating significant accuracy gain with a justifiable and manageable computational overhead.

\subsection{Integrated System Performance on Hardware}
\begin{figure*}[h!]
    \centering
    \includegraphics[width=\textwidth]{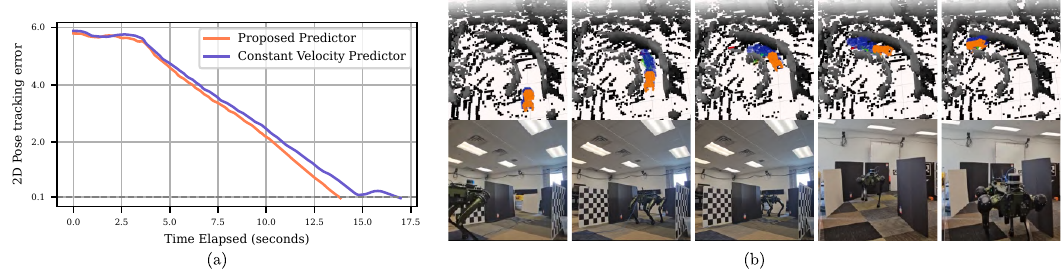}
    \vspace*{-8mm}
    \caption{\textbf{Real-Robot Experiments.} (a) Pose tracking error for a representative trial. \ac{MPPI} with our predictor (orange) converges faster and more smoothly than with the CV baseline (blue), which exhibits oscillations due to unmodeled dynamics. (b) The Vision 60 robot successfully navigating a narrow passage using our full framework, which enables limb-aware, collision-free planning.}
    \label{fig:experiments}
    \vspace*{-5mm}
\end{figure*}
To demonstrate practical effectiveness, we integrate the learned predictor into an \ac{MPPI} planner to perform navigation tasks as described in Sec.~\ref{sec:mppi_sec}.

\subsubsection{Experimental Setup}
Experiments are conducted on the Vision 60 hardware. The robot's state is estimated via a contact-aided InEKF \cite{Hartley2020IJRR}. An onboard VLP-16 LiDAR is used to build a voxel map \cite{Hornung2013AURO} of the environment for collision checking. The \ac{MPPI} planner operates at $25$~Hz, using $N=1000$ samples over a $T=200$ step ($4$~s) horizon. With our predictor, a full planning cycle (including prediction and collision checks) takes approximately $35$~ms on an Nvidia RTX T1000 GPU, well within the $40$~ms real-time budget. For all trials, the \ac{MPPI} constants are held fixed: $d_\mathrm{switch}=0.5$, $s_\mathrm{switch} = 0.5$, $\mathrm{w}_\mathrm{c}=500$, $\mathrm{w}_\mathrm{g}=10$, and $\mathrm{w}_\mathrm{u}=0.01$. We compare the performance of \ac{MPPI} using our predictor against the CV baseline over 20 trials for each task.

\begin{table}[h]
\centering
\setlength{\extrarowheight}{3pt}
\caption{MPPI PERFORMANCE OVER 20 TRIALS}
\label{tab:experiments}
\begin{tabular}{lcc}
\hline
\textbf{Predictor Model} & \textbf{Time to Track (s)} & \textbf{Failure Rate (\%)} \\ \hline
Constant Velocity (Baseline) & $2.57 \pm 0.56$ & $40\%$ \\
\textbf{Proposed Predictor (Ours)} & $\mathbf{2.45 \pm 0.37}$ & $\mathbf{15\%}$ \\ \hline
\vspace*{-5mm}
\end{tabular}
\end{table}

\subsubsection{Goal-Pose Tracking Performance}
We first evaluate tracking a goal pose of $[5.0~\mathrm{m}, 2.5~\mathrm{m}, 1.57~\mathrm{rad}]$ in an open area. As shown in Fig.~\ref{fig:experiments}(a) and Table~\ref{tab:experiments}, our predictor enables faster and more consistent tracking. The CV baseline suffers from poorer performance because it cannot account for complex behaviors, such as the low-level controller's tendency to enter a "standing mode" at very low velocities, which our model learns from data. This leads to oscillations in the error plot for the baseline as it nears the goal.

\subsubsection{Collision Avoidance in a Narrow Passage}
Next, we task the robot with navigating a narrow passage (Fig.~\ref{fig:experiments}(b)), where the ability to predict the full-body configuration $\mathbf z_k$, including limb motion, is critical. Our predictor allows for limb-level collision checking with a robot occupancy predictor (see Fig.~\ref{fig:occupancy_predictor}), whereas the CV baseline is limited to collision checks using a static, zero-configuration body model. For a fair comparison, robot occupancy predictions were not inflated for either model. The results in Table~\ref{tab:experiments} show a huge improvement in failure rates from $40\%$ to $15\%$. Failures with the CV model were often catastrophic collisions, while failures with our model were typically minor scrapes, highlighting its ability to generate safer, more aware motion plans.

\subsubsection{Navigation Over Obstacles}
\begin{figure}[h]
    \centering
    \includegraphics{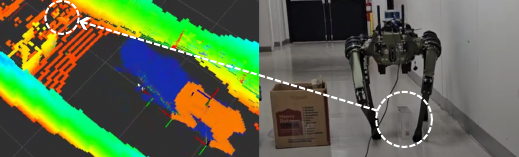}
    \vspace*{-8mm}
    \caption{Limb-aware predictions enables collision-free navigation over small objects.}
    \label{fig:navigation_over_box}
\end{figure}
Finally, we tested the full system's ability to plan collision-free paths over small objects. As shown in Fig.~\ref{fig:overview} and Fig.~\ref{fig:navigation_over_box}, our predictor's ability to forecast future limb positions enables the planner to successfully navigate over clutter. We observed that the efficacy of the plan is highly dependent on the quality of the onboard map and the drift of the state estimator \cite{Hartley2020IJRR}.

\section{Conclusions and Future Work}
In this work, we introduced a learning-based framework for full-body motion prediction to enable efficient, safe, collision-aware navigation for legged robots. Our approach features a decoupled observer-predictor neural architecture, where a neural observer with provable \ac{UUB} guarantees provides a reliable latent state estimate to a computationally efficient recurrent predictor. This theoretical guarantee is critical, as it also facilitates stable end-to-end training of the proposed framework. 
We validated our method through hardware experiments on a Vision 60 quadruped. 
The results demonstrate that when integrated into an MPPI planner, our model significantly outperforms a kinematic baseline in both goal-pose tracking and collision avoidance. 
A key limitation of the current work is its focus on flat-ground locomotion; therefore, future work will be directed toward extending this framework to handle complex and uneven terrain.

\appendices
\section{Proof of Lem.~\ref{lem:relu_lips}} 
\label{sec:proof_relu_lem}

For each layer of MLP 
$g^n(\mathbf x)= \mathbf W^n \mathbf x+\mathbf b^n$, 
$ \|g^n(\mathbf x)-g^n(\mathbf y)\|_2=\|\mathbf W^n(\mathbf x-\mathbf y)\|_2\le \|\mathbf W^n\|_2\|\mathbf x-\mathbf y\|_2$, 
we thus have  $L_{g^n}=\|\mathbf W^n\|_2$ (note that biases do not affect Lipschitz constants). 
The ReLU is 1-Lipschitz since, coordinate-wise, $|\phi(\mathbf u_i)-\phi(\mathbf v_i)|\le |\mathbf u_i-\mathbf v_i|$, hence $\|\phi(\mathbf u)-\phi(\mathbf v)\|_2\le \|\mathbf u-\mathbf v\|_2$.  
Writing $g=g^{N+1}\circ \phi \circ g^{N}\circ \cdots \circ \phi \circ g^{1}$ and noting that Lipschitz constants multiply under composition: 
$L_{f\circ g}\le L_fL_g$
we obtain:
\begin{equation*}
L_g\le \|\mathbf W^{N+1}\|_2\prod_{n=1}^{N}\bigl(1\cdot \|\mathbf W^{n}\|_2\bigr) = \prod_{n=1}^{N+1}\|\mathbf W^{n}\|_2 .
\end{equation*}

\section{Proof of Thm.~\ref{thm:observer_stability_contraction}}
\label{sec:proof_uub_thm}

The dynamics of the estimation error $\mathbf e_k = \mathbf x_k - \hat{\mathbf x}_k$ are derived by subtracting the observer dynamics from the true system dynamics:
\begin{align*}
    \mathbf e_{k+1} &= (\mathbf A - \mathbf K \mathbf C_y) \mathbf e_k + \big(g(\mathbf x_k, \mathbf u_k) - g(\hat{\mathbf x}_k, \mathbf u_k)\big) + \bm\epsilon_k.
\end{align*}
Letting $\mathbf A_c = \mathbf A - \mathbf K \mathbf C_y$, we take the 2-norm and apply the triangle inequality:
\begin{align*}
    \|\mathbf e_{k+1}\|_2 &\le \|\mathbf A_c \mathbf e_k\|_2 + \|g(\mathbf x_k, \mathbf u_k) - g(\hat{\mathbf x}_k, \mathbf u_k)\|_2 + \|\bm\epsilon_k\|_2 \\
    &\leq \|\mathbf A_c\|_2\|\mathbf e_k\|_2 + L_g \|\mathbf e_k\|_2 + \bm\epsilon_{\max} \\
    &= (\|\mathbf A_c\|_2 + L_g)\|\mathbf e_k\|_2 + \bm\epsilon_{\max}.
\end{align*}
With $\rho = \|\mathbf A_c\|_2 + L_g < 1$, the error norm dynamics are bounded by the scalar sequence $\|\mathbf e_{k+1}\|_2 \leq \rho \|\mathbf e_k\|_2 + \bm\epsilon_{\max}$. By recursive application, this inequality shows the sequence is bounded, and taking the limit superior yields the ultimate bound for the state error:
\begin{equation*}
    \limsup_{k \to \infty} \|\mathbf e_k\|_2 \leq \frac{\epsilon_{\max}}{1 - \rho}.
\end{equation*}

We now examine the output error, $\mathbf e_{y,k} = \mathbf y_k - \hat{\mathbf y}_k$. 
By definition of the output map, we have $\mathbf e_{y,k} = \mathbf C_y \mathbf x_k - \mathbf C_y \hat{\mathbf x}_k = \mathbf C_y \mathbf e_k$. By taking the 2-norm we have:
\begin{equation*}
    \|\mathbf e_{y,k}\|_2 = \|\mathbf C_y \mathbf e_k\|_2 \leq \|\mathbf C_y\|_2 \|\mathbf e_k\|_2.
\end{equation*}
With established ultimate bound for $\|\mathbf e_k\|_2$, we can directly find the ultimate bound for the output error:
\begin{equation*}
    \limsup_{k \to \infty} \|\mathbf e_{y,k}\|_2 \leq \|\mathbf C_y\|_2 \left( \limsup_{k \to \infty} \|\mathbf e_k\|_2 \right) \leq \|\mathbf C_y\|_2 \frac{\epsilon_{\max}}{1 - \rho}.
\end{equation*}

\bibliographystyle{IEEEtran}
\bibliography{IEEEabrv,biblo}

\end{document}